# Merging Language and Domain Specific Models: The Impact on Technical Vocabulary Acquisition

Thibault Rousset[1], Taisei Kakibuchi[2], Yusuke Sasaki[2], and Yoshihide Nomura[2]

[1]School of Computer Science, McGill University
thibault.rousset@mail.mcgill.ca
[2]Fujitsu Research Ltd.

## ABSTRACT

*Advancements in Natural Language Processing have enabled specialized language models, but integrating domain-specific knowledge into general-purpose models in multilingual settings remains challenging, particularly for technical vocabulary. This paper explores cross-lingual knowledge transfer in model merging, examining how combining a general-purpose language model with a domain-specific model affects technical jargon comprehension. The objective is to evaluate the effectiveness of merging techniques in enhancing domain-specific proficiency while preserving general language understanding. Our study analyzes different merging strategies and their impact on specialized terminology retention. A quantitative evaluation compares the merged model's performance against its constituent models, offering insights into the strengths and limitations of various approaches. The results demonstrate the potential of model merging for domain adaptation while highlighting challenges in cross-lingual knowledge transfer. These findings provide valuable guidance for optimizing model merging techniques in specialized NLP applications.*

## 1. INTRODUCTION

The rapid advancement of Natural Language Processing (NLP) has led to the widespread adoption of Large Language Models (LLMs) across diverse applications. While these models exhibit remarkable versatility [1,2], their effectiveness in specialized domains remains limited due to insufficient exposure to domain-specific knowledge during training. This issue becomes particularly pronounced in fields such as medicine, law, and engineering, where precise understanding and accurate generation of technical terminology are essential. Even minor misunderstandings in these domains can lead to significant misinterpretations, impacting decision-making and real-world applications. Therefore, developing methods to effectively incorporate domain-specific knowledge into LLMs is vital for enhancing their applicability and reliability in specialized contexts.

One promising approach to addressing this limitation is model merging, which integrates the strengths of multiple LLMs to enhance domain adaptation. Model merging presents a cost-effective alternative to full-scale retraining or fine-tuning, allowing the integration of new knowledge without requiring large amounts of additional data or computational resources. However, the extent to which model merging facilitates domain-specific knowledge integration, particularly in multilingual settings, remains an open question. This limitation is particularly problematic for applications that require precise understanding and generation of technical language. An accurate interpretation of terms and concepts is essential in these fields, as even minor misunderstandings can lead to significant errors or miscommunications.

This study explores the potential of model merging for cross-lingual knowledge transfer, with a particular focus on integrating domain-specific technical vocabulary. The primary challenge lies in ensuring effective knowledge transfer without interference so that newly acquired domain-specific information enhances the model's proficiency while preserving its general linguistic capabilities. Another key issue is whether merging enables the model to retain and accurately utilize domain-specific terminology across different languages, maintaining both contextual meaning and usability in a multilingual setting. To investigate this, we conduct a comprehensive experiment, merging a general-purpose Japanese-specific model with an English medical domain-specific model and assessing various merging strategies. Through quantitative analysis, we evaluate the effectiveness of different approaches in transferring domain-specific terminology knowledge and improving the model's ability to understand technical language, particularly medical jargon. By comparing the performance of merged models with their original components, we aim to determine the extent to which merging allows models to leverage both general and specialized knowledge across languages. Our findings provide empirical insights into the complexities of model merging for domain adaptation and cross-lingual knowledge transfer, offering guidance on optimizing merging strategies for NLP applications. Ultimately, we seek to enhance the ability of language models to handle domain-specific terminology, bridging the gap between general and specialized language capabilities in multilingual settings and advancing the development of more versatile NLP models for specialized applications.

## 2. PAPER ORGANIZATION

This paper is structured to provide a comprehensive analysis of model merging for cross-lingual knowledge transfer and its impact on technical vocabulary acquisition. **Section 3** presents a review of related work, discussing existing research on model merging, domain adaptation, and cross-lingual knowledge transfer. **Section 4** describes the experimental setup, including details on the selected models, dataset preparation, merging techniques, and evaluation methodologies. **Section 5** provides an in-depth analysis of the results, comparing the performance of merged models with their constituent models and examining the strengths and limitations of different merging strategies. **Section 6** concludes the paper by summarizing key findings, highlighting current challenges, and proposing potential directions for future research in improving cross-lingual domain adaptation through model merging.

## 3. RELATED WORK

Large Language Models (LLMs) have seen significant advancements in recent years, with researchers exploring various techniques to enhance their adaptability and performance across different domains and languages. Two key areas of study in this context are model merging for domain adaptation and cross-lingual knowledge transfer. This section reviews relevant research in these areas, highlighting key methodologies, challenges, and ongoing developments.

### 3.1. Model Merging for LLM Domain Adaptation

Model merging methods, widely used across diverse fields within NLP, are increasingly employed for LLM domain adaptation [3]. This approach involves combining the strengths of multiple models – often a general-purpose LLM with one or more domain-specific models – to enhance performance in a targeted domain. The aim is to leverage the broad knowledge base of the general LLM while incorporating the specialized expertise of the domain models, creating a hybrid system that surpasses the capabilities of its individual components. However, effective model merging requires careful consideration of model compatibility, potential knowledge interference, and computational efficiency. Ongoing research focuses on developing optimal merging strategies and addressing the complexities of integrating diverse knowledge sources without compromising overall model performance.

Model merging offers advantages over other domain adaptation techniques like finetuning [4] and continual learning [5,6], particularly regarding data requirements and computational costs. Fine-tuning often necessitates substantial labeled domain-specific data, which may be scarce and can be computationally expensive, while continual learning can be susceptible to catastrophic forgetting [7]. Model merging, in contrast, leverages pretrained models, reducing the need for extensive retraining and minimizing computational overhead. However, the choice of merging method can introduce constraints; some methods may require models of similar size and architecture, potentially limiting the flexibility of model selection and hindering the benefits of combining models with complementary strengths. Current research aims to overcome these limitations by developing more flexible and efficient model merging strategies for optimal LLM domain adaptation.

### 3.2. Cross-lingual Knowledge Transfer

Cross-lingual knowledge transfer refers to the ability of a model to leverage knowledge learned in one language to enhance performance in another, which is crucial for developing models that can operate effectively in various languages by transferring insights from high-resource languages. As multilingual LLMs become more prevalent and performant, understanding the mechanisms of knowledge sharing and transfer across languages has become increasingly important. This field has gained significant attention with the rise of massively multilingual LLMs, which have opened new avenues for natural language processing by enabling the transfer of knowledge across linguistic boundaries [8,9,10].

The primary objective of cross-lingual knowledge transfer is to utilize information learned in one language to improve performance in another, thereby benefiting languages with limited training data. Recent trends in this field focus on understanding the mechanisms that facilitate such knowledge transfer, with researchers exploring how multilingual models can share and transfer knowledge across languages [11,12]. There is a growing interest in developing evaluation frameworks that accurately assess cross-lingual capabilities [13], providing insights into the strengths and limitations of current models.

Efforts are also being made to address challenges such as language-specific biases and cultural nuances that can hinder effective knowledge transfer [14,15,16]. By refining models to better handle these complexities, researchers aim to create more robust and versatile multilingual systems. The objective is to enhance the adaptability and performance of LLMs across diverse linguistic contexts, ensuring they can meet the demands of global applications. As the field progresses, it continues to push the boundaries of what is possible in multilingual natural language processing, striving for seamless integration and transfer of knowledge across languages.

### 4. EXPERIMENT

To assess the effectiveness of model merging for cross-lingual domain adaptation, we conduct experiments evaluating various merging methods applied to a Japanese general-purpose LLM and an English medical domain-specific model. The goal is to determine to what extent domain-specific knowledge can be transferred across languages through merging, particularly in defining technical medical terms. This section details our methodology, dataset preparation, evaluation criteria, experimental settings, and baseline comparisons, providing a comprehensive framework for analyzing the impact of model merging on linguistic and domain-specific proficiency.

## 4.1. Methodology

This study aims to evaluate the effectiveness of various model merging methods in enhancing a Japanese general-purpose LLM's ability to understand and generate definitions of technical medical terms by merging it with an English medical domain-specific model. Both constituent models are fine-tuned from the same pre-trained Llama 3 [1] model, and a wide range of merging methods is explored. The overarching goal is to evaluate to what extent such methods allow effective knowledge transfer of domain-specific terminology across languages. The methodology involves several key steps that will be detailed later:

**Model Preparation** We select a Japanese general-purpose LLM and an English medical domain-specific LLM derived from the same pre-trained model to ensure coherence and avoid merging artifacts. These models serve as the foundational components for our merging experiments. The Japanese model provides the linguistic base, while the English model offers specialized medical knowledge. The objective is then to incorporate the English technical knowledge from the domain-specific model into the Japanese model, using the LLMs' cross-lingual capacities and knowledge transfer mechanisms of the merging.

**Merging Methods** We apply six different merging methods to combine the models. Each method is designed to integrate the strengths of both models, with specific parameters adjusted to optimize the knowledge transfer. The merging process involves aligning the models' representations and combining their knowledge bases to create a unified model capable of handling both general and technical data in both languages.

**Dataset Preparation** A curated list of technical medical terms is compiled from the Systematized Nomenclature of Medicine (SNOMED) [17], ensuring that only relevant and specialized vocabulary is included. These terms are then translated into Japanese using GPT-4o [2] to maintain consistency with the evaluation process and minimize translation bias.

**Definition Generation and Evaluation** The merged models, along with the baseline Japanese and English models, are tasked with generating definitions for each of the curated technical medical terms. Two judge LLMs evaluate the accuracy of these definitions, GPT-4o [2] and Gemini 1.5 Pro [18], which score the accuracy of the definitions out of ten. The evaluation context includes a baseline English definition generated by the English expert model, providing a reference point.

**Analysis and Comparison** The performance of each merged model is evaluated in comparison to its constituent models using the scores provided by the LLMs. Additionally, to assess the models' overall knowledge and capabilities, we evaluate them against established Japanese language proficiency benchmarks and medical terminology benchmarks. This dual evaluation approach allows us to compare the models' relative performance in generating accurate definitions and their absolute knowledge levels in both language and domain-specific contexts. By analyzing these results, we identify strategies that most effectively enhance the Japanese model's proficiency in technical language comprehension. At the same time, this evaluation clarifies potential limitations in transferring specialized terminology. Overall, our findings provide insights into the strengths and weaknesses of model merging as a crosslinguistic knowledge transfer mechanism, offering a clearer picture of how it can influence the model's broader knowledge acquisition.

This comprehensive methodology provides a structured framework for evaluating the impact of model merging on domain-specific language capabilities, contributing to the development of more versatile and capable language models.

## 4.2. Dataset and Evaluation

**Vocabulary Dataset and Evaluation** The vocabulary dataset was curated from the most recent data of the Systematized Nomenclature of Medicine [17] and refined by retaining only terms with a frequency of 1 or 0 in the widely used Brown Corpus [19]. This approach ensures that the dataset comprises only specialized and rare technical vocabulary, which is infrequently encountered in general literature and largely absent from the training data of most general-purpose pre-trained language models. To further enhance precision, we limited the dataset to nouns and adjectives, thereby minimizing translation artifacts and avoiding potential ambiguities in Japanese, where part-of-speech tagging significantly differs from that of English. The final curated dataset includes 1,782 technical terms drawn from conventional medical terminology. These terms were then translated into Japanese using GPT-4o for integration with Japanese language models.

Our evaluation method measures model performance by assessing their ability to define technical terms in the target language. Each model receives a term and an instruction prompt, guiding them to generate a definition that reflects their understanding of domain-specific terminology. The English expert model is tested using English terms, while Japanese terms are used for others, ensuring that evaluation remains within the appropriate linguistic context. The generated definitions are then reviewed by judge models—GPT-4o and Gemini 1.5 Pro—with GPT-4o as the primary evaluator to mitigate translation bias, considering its role in term translation. The judges receive the term in both English and Japanese, along with the English expert model definition as a reference, and assign accuracy scores on a ten-point scale. We then analyze the score distributions and statistical trends to compare model performance and examine cross-lingual knowledge transfer dynamics.

**Benchmark Assessment** In addition to term definition evaluation, we assess our models on three distinct benchmarks to measure high-level linguistic and technical knowledge and identify potential cross-lingual knowledge transfer. By comparing performance against the base models, we verify that fundamental linguistic and domain expertise is preserved while evaluating the extent of knowledge integration and transfer across languages. These benchmarks evaluate general Japanese linguistic proficiency to ensure fluency in the target language, as well as medical domain knowledge in both English and Japanese to assess technical depth. The three publicly available benchmarks used in our evaluation are:

> **JMedBench** [20] – JMedBench is a comprehensive benchmark designed for evaluating Japanese biomedical large language models (LLMs), developed by the University of Tokyo and the National Institute of Informatics. It encompasses 20 datasets across five key tasks: multi-choice question answering (MCQA), named entity recognition (NER), machine translation (MT), document classification (DC), and semantic text similarity (STS). This benchmark integrates both human-created and translated datasets to provide a robust evaluation framework. In our evaluation, we focus on the multi-choice question answering (MCQA), document classification (DC), and semantic text similarity (STS) tasks, leveraging JMedBench's extensive resources to assess domain-specific expertise in Japanese.

> **PubMedQA** [21] – A benchmark for evaluating LLMs in biomedical question answering using PubMed abstracts. It consists of expert-labeled yes/no/maybe answers to research questions, testing models' ability to understand and reason over medical literature in English.

> **Japanese LLM Leaderboard** – A benchmark evaluating Japanese language proficiency across diverse NLP tasks. It includes question answering (JAQKET, JSQuAD [22]), text summarization (XL-Sum [23]), pronoun resolution (XWinograd [24,25]), and JCommonsenseQA [22]), providing a comprehensive assessment of linguistic and reasoning abilities in Japanese.

Our evaluation method combines specialized term definition tasks with established benchmarks to assess both linguistic proficiency and technical expertise across languages. By incorporating diverse evaluation metrics, we ensure a comprehensive understanding of model performance while also verifying that cross-lingual knowledge transfer does not compromise domain-specific knowledge retention and vice versa. This approach allows us to rigorously analyze model capabilities and the effectiveness of integrating specialized vocabulary in different linguistic contexts.

### 4.3. Experimental Settings

For our experiments, we selected two base models, each with 8B parameters: Suzume [26] as the general-purpose Japanese model, and ContactDoctor-8B [27], as the English medical domain-specific model. Both models were fine-tuned from the same Llama-3-Instruct [28] pre-trained base model. These models were chosen for their open-source availability, relatively low computational requirements, and strong performance relative to their size.

We applied six different model merging methods to integrate knowledge across the base models: TIES [29], Task Arithmetic [30], SLERP, Linear [31], DARE TIES [32], and an Evolutionary variant of DARE TIES [33]. When necessary, the Japanese model was used as the base model. The merging process was carried out using the MergeKit framework [34]. Although we conducted hyperparameter tuning for the merging methods, the results showed minimal impact on downstream performance. As a result, we selected the median values for the hyperparameters across all configurations. The Evolutionary Merging optimization was performed on the PubMedQA and Japanese LLM Leaderboard benchmarks, as they align with the downstream tasks we aimed to optimize.

For evaluation, we employed GPT-4o from the GPT suite [2] and Gemini 1.5 Pro from the Gemini series [18] as judge models, chosen for their proven performance in multilingual settings and widespread use in evaluation tasks. Definitions were generated in a zero-shot setting, without sampling, and with a token limit of 256. Benchmark evaluation utilized the LM Evaluation Harness framework [35]; PubMedQA and XWinograd, as well as all JMedBench tasks, were evaluated in a zero-shot setting, while the remaining Japanese LLM Leaderboard benchmarks were evaluated using a few-shot approach. All inference and evaluation experiments were carried out on an A100 GPU.

### 4.4. Baselines

For evaluation, we establish several baselines to assess the effectiveness of the merging methods. To establish a consistent evaluation framework, the English expert model was prompted twice for each term: once to generate a reference definition, which served as context for the judge models, and a second time to generate a definition for baseline evaluation. This ensured that the baseline model's performance was assessed using the same process and criteria as the merged models, providing a fair and comparable standard for performance analysis. For benchmark evaluation, constituent models serve as baselines, enabling us to analyze whether knowledge has been transferred, retained, or forgotten through the merging process.

## 5. RESULTS AND ANALYSIS

Our study reveals contrasting results for model merging effectiveness. While merged models demonstrate satisfactory performance on benchmarks requiring monolingual knowledge transfer, they perform poorly on those necessitating cross-lingual knowledge transfer and on vocabulary acquisition evaluation. This discrepancy highlights that although general knowledge transfer within a single language is effective, the integration of domain-specific terminology across languages is significantly weaker. A notable performance gap persists, particularly for tasks requiring cross-lingual knowledge transfer, suggesting that this shortfall in transferring technical vocabulary is a key contributor to the remaining limitations, especially on Japanese medical benchmarks. These results underscore the need for improved merging strategies to fully leverage cross-lingual capabilities.

Table 1. Comparison of Model Performance based on Definition Evaluation Scores

| Model | GPT Scores | | | Gemini Scores | | |
|---|---|---|---|---|---|---|
| | Median | Mean | Std | Median | Mean | Std |
| Baseline | 10 | 9.48 | 1.66 | 10 | 9.30 | 2.07 |
| Base Japanese | 6.0 | 5.46 | 3.13 | 6.0 | 5.14 | 3.29 |
| SLERP | 6.0 | 5.64 | 3.15 | 7.0 | 5.32 | 3.36 |
| TIES | 6.0 | 5.33 | 3.17 | 6.0 | 5.18 | 3.38 |
| Linear | 6.0 | 5.33 | 3.08 | 6.0 | 5.09 | 3.45 |
| Task Arithmetic | 5.0 | 5.14 | 3.13 | 6.0 | 4.95 | 3.38 |
| DARE TIES | 5.0 | 5.20 | 3.14 | 5.0 | 4.90 | 3.38 |
| Evo. DARE TIES | 5.0 | 5.13 | 3.14 | 4.0 | 4.88 | 3.36 |

Table 1 and Figure 1 present the general results of our definition evaluation experiment. The baseline model achieves near-perfect scores, which confirms that our evaluation method using the judge models is robust and unbiased. However, when examining the merged models, there is minimal change in performance compared to the base non-expert Japanese model. This is evident both in the statistical metrics and in the score distributions, with most models showing a slight decline in performance, except for SLERP, which demonstrates a modest improvement but negligible when compared to the baseline.

The benchmark results in Table 2 illustrate that while knowledge transfer within a single language is evident, as demonstrated by the performance on PubMedQA and the Japanese Leaderboard, where merged models variably achieve good results compared to the baselines—sometimes even outperforming them in certain tasks—this success does not extend to cross-lingual knowledge transfer, as demonstrated by JMedBench results, where merged models generally fail to surpass the base models and, in most cases, perform significantly worse than the non-expert model. When these findings are juxtaposed with the vocabulary acquisition results, it becomes clear that the transfer of higher-level knowledge is impeded substantially by inadequate cross-lingual knowledge transfer at the technical terminology level.

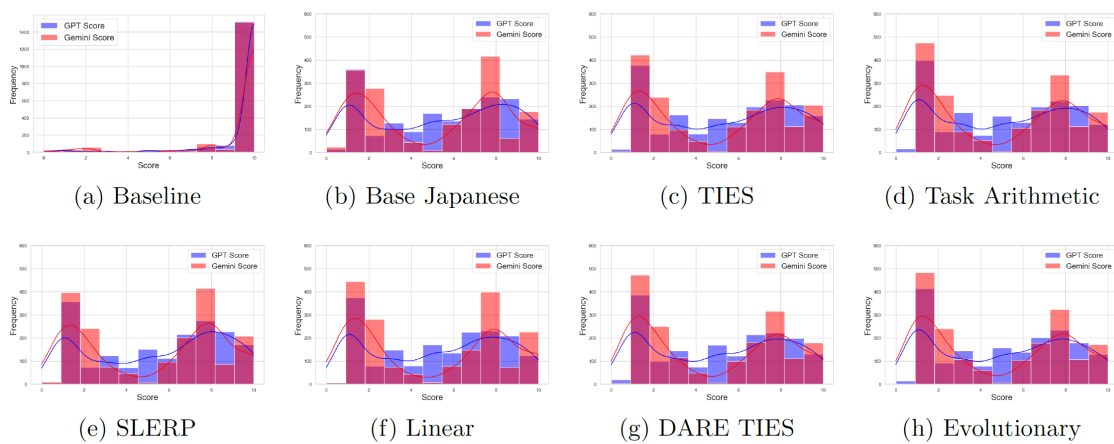

(a) Baseline  (b) Base Japanese  (c) TIES  (d) Task Arithmetic
(e) SLERP  (f) Linear  (g) DARE TIES  (h) Evolutionary

Figure 1. Histograms of GPT and Gemini Scores for definition evaluation across baseline, base Japanese, and merged models.

Table 2. Benchmark Performance Summary. Results in **bold** (<u>underline</u>) indicate best (worst) performance. E.M. denotes Exact Match, and Acc. denotes accuracy.

| Benchmark | PubMedQA | JAQKET v2 | JComQA | JSQuAD | XWinograd | XL-Sum | MCQA | DC | STS |
|---|---|---|---|---|---|---|---|---|---|
| Metric | Acc. | E.M | Acc. | E.M | Acc. | ROUGE | Acc. | Acc. | Pearson |
| Baseline EN | 0.804 | <u>0.604</u> | 0.696 | 0.582 | 0.715 | <u>0.018</u> | 0.288 | 0.416 | <u>0.351</u> |
| Baseline JP | <u>0.734</u> | **0.762** | **0.788** | 0.642 | **0.738** | **0.065** | 0.290 | **0.424** | **0.559** |
| SLERP | 0.786 | 0.739 | 0.784 | **0.655** | 0.730 | 0.037 | 0.290 | 0.401 | 0.516 |
| TIES | 0.800 | 0.611 | 0.734 | 0.609 | 0.727 | 0.020 | 0.292 | 0.416 | 0.433 |
| Linear | 0.786 | 0.738 | 0.784 | 0.654 | 0.727 | 0.041 | 0.289 | <u>0.397</u> | 0.511 |
| Task Arithmetic | **0.806** | 0.610 | 0.699 | 0.582 | 0.714 | 0.019 | 0.288 | 0.416 | 0.385 |
| DARE TIES | 0.798 | 0.605 | <u>0.687</u> | <u>0.580</u> | <u>0.706</u> | 0.019 | **0.336** | 0.408 | 0.384 |
| Evo. DARE TIES | **0.806** | 0.621 | 0.715 | 0.584 | 0.712 | 0.026 | <u>0.285</u> | 0.412 | 0.405 |

The performance of the Evolutionary model presents a compelling case of concordance between its results on vocabulary acquisition and the JMedBench, both of which are notably poor. This alignment suggests a likely causal relationship, indicating that deficiencies in technical vocabulary acquisition may directly affect the effectiveness of cross-lingual knowledge transfer at higher levels. Interestingly, the model achieves satisfactory results on other benchmarks, even surpassing the expert model on PubMedQA. This suggests that model merging can be an effective strategy for knowledge transfer in specialized domains. However, it also highlights that while effective knowledge transfer at a monolingual level and strong performance on downstream tasks inherited from constituent models are evident, they do not necessarily translate to, and might even hinder, cross-lingual knowledge transfer.

## 6. CONCLUSION

In this study, we explored the potential of model merging methods to enhance the integration of technical vocabulary in language models, particularly focusing on cross-lingual knowledge transfer. Our findings reveal that while model merging can facilitate knowledge transfer at a mono-lingual level, as evidenced by satisfactory performance on general benchmarks, it struggles with the effective acquisition and integration of technical terminology across languages. The merged models' performance remained similar to that of the non-expert Japanese model, with a tendency towards decline, highlighting the challenges of incorporating specialized vocabulary.

The results underscore the complexity of achieving effective cross-lingual knowledge transfer, particularly in domains requiring precise technical language comprehension. The observed performance suggests that current merging methods may introduce complexities that hinder the integration of domain-specific terminology. Despite these challenges, the study provides valuable insights into the strengths and limitations of model merging, offering a foundation for future research aimed at developing more sophisticated methods for domain adaptation and cross-lingual knowledge transfer.

Research should then focus on refining existing merging methods to better handle technical vocabulary and explore alternative strategies that enhance the integration of specialized knowledge without compromising the models' general capabilities. By addressing these challenges, we can advance the development of more versatile and capable language models suitable for specialized applications across diverse linguistic contexts.

While this study provides valuable insights into the potential of model merging for cross-lingual technical vocabulary acquisition, several limitations must be acknowledged. Firstly, the reliance on judge LLMs for evaluating definition accuracy introduces a degree of uncertainty, as these models may not fully capture the nuances of human judgment. Secondly, the study's focus on the Japanese-English language pair and the medical domain may limit the generalizability of the findings. The significant linguistic differences between these languages likely influence the results, and effectiveness could vary with other language pairs or domains. Thirdly, the investigation is limited to six specific model merging methods, and exploring alternative approaches could reveal more effective strategies. Finally, the minimal impact of hyperparameter tuning suggests robustness, but further optimization could potentially yield improved results. Addressing these limitations is crucial for future research to achieve a comprehensive and detailed understanding of these mechanisms and to develop new, more performant techniques.